\documentclass[runningheads]{llncs}

\usepackage{graphicx}
\usepackage{marvosym}   % \Letter -- the corresponding-author mark
\usepackage{amsmath}
\usepackage{amssymb}
\usepackage{booktabs}
\usepackage{multirow}
\usepackage{xcolor}
\usepackage[hidelinks]{hyperref}
\usepackage{siunitx}
\usepackage{pgfplots}
\pgfplotsset{compat=1.18}
\usepgfplotslibrary{groupplots}
\usepackage{algorithm}
\usepackage[noend]{algpseudocode}   % no "end for" line; the loop body is shown by indentation

\newcommand{\thetaS}{\theta}          % task adapters (shared, fixed at inference)
\newcommand{\phiF}{\varphi}           % case adapters (reset per case, updated at test time)
\newcommand{\dice}{\mathrm{Dice}}

\begin{document}

\title{LeCor: Learning to Be Corrected by Meta-Learned Test-Time Training\\
       for Interactive 3D Lung-Tumour Segmentation}
\titlerunning{LeCor: Learning to Be Corrected}

% TODO before submission: confirm affiliations. All five authors are given here as Johns
% Hopkins University per the corresponding author's instruction; note that Yike Guo is
% publicly listed at HKUST, so please double-check before you submit.
\author{Yi~Luo\inst{1} \and
        Yike~Guo\inst{2} \and
        Wenxuan~Li\inst{3} \\
        Zongwei~Zhou\inst{3} \and
        Rui~Zhang\inst{4}\textsuperscript{(\Letter)} \and
        Kai~Ding\inst{2}\textsuperscript{(\Letter)}}

\authorrunning{Y. Luo et al.}

\institute{Department of Biomedical Engineering, Johns Hopkins University \and
           Department of Radiation Oncology and Molecular Radiation Sciences, Johns Hopkins University \and
           Department of Computer Science, Johns Hopkins University \and
           Department of Surgery, University of Minnesota\\
           \email{ruizhang@umn.edu, kai@jhu.edu}}

\maketitle

% =====================================================================================
\begin{abstract}
Delineating lung tumours on computed tomography (CT) takes a considerable share of the
time spent on radiotherapy planning, and a contour proposed by a model can be refined
interactively by the clinician. Promptable foundation models such as SAM~3 support this
workflow by writing each correction into a session memory that conditions the remaining
slices, while the model weights stay fixed. On 690 test cases from five public CT cohorts, fine-tuning SAM~3 on lung tumours raises
the Dice obtained from a single point prompt from 0.298 to 0.757, and seven rounds of
corrections raise it further to 0.765, but under memory conditioning alone the accuracy
on slices the annotator has not touched stops improving after six rounds. We therefore treat each correction as a training
signal and propose LeCor, which performs test-time training on a small set of case
adapters that are reset for every case and meta-learned such that a single gradient step
driven by a click improves the slices that were not clicked. On the
133 test cases that span at least eight slices, LeCor raises the Dice reached after seven
correction rounds from 0.787 with the fine-tuned model to
0.827, reduces the number of cases that never reach a Dice of 0.80 from 47 to 27, and
reaches in three correction rounds the accuracy that the fine-tuned model attains in
seven.
\keywords{Interactive segmentation; Test-time training; Meta-learning; Foundation models;
Lung cancer; Radiotherapy planning}
\end{abstract}

% =====================================================================================
\section{Introduction}
\label{sec:intro}

Radiotherapy for lung cancer requires the gross tumour volume to be delineated on the
planning CT. Manual delineation takes roughly 16 to 22 minutes per case, and when eleven
radiation oncologists delineated the same tumours, the three dimensional variation
between them was about 1\,cm at one standard deviation
\cite{steenbakkers2005observer,steenbakkers2006reduction,trimpl2024interactive}. Fully
automatic segmentation with the reference method nnU-Net \cite{isensee2021nnunet} still
leaves every contour to be reviewed and edited by a clinician before it is used. Clinical
workflows have therefore converged on an interactive workflow in
which the model proposes a contour and the clinician refines it. For thoracic organs at
risk and for lung tumours, editing model proposals was faster and more consistent than
manual delineation
\cite{lustberg2018clinical,vaassen2020evaluation,trimpl2024interactive,han2024ai}, and
starting from an automatic contour reduced the variability between observers in most of
the intervention studies.\ \cite{vinod2016interobserver}. This
work addresses interactive segmentation in which the clinician supplies corrections as
clicks and the model revises its prediction after each one.

Promptable foundation models have made this workflow inexpensive. The Segment Anything
Model (SAM) \cite{kirillov2023sam} segments an object from a point or box prompt, SAM~2
\cite{ravi2024sam2} extends it to video by carrying a memory of previously segmented
frames, and SAM~3 \cite{sam3} builds on the same memory architecture. Treating the slice
axis of a CT volume as time turns volumetric segmentation into a tracking problem, and
this is how these models are applied to 3D medical images
\cite{shen2024sam2interactive3d,zhu2024medicalsam2,ma2025medsam2}. The click is encoded into the
session memory, and every slice decoded afterwards attends to that memory, so that a
single correction can improve slices that the annotator never touched. However, the weights of the model stay fixed for the whole session. Each correction is consumed as a
conditioning input; its contribution is bounded by what the memory can carry and is not
retained in the parameters of the model.

Test-time training offers a different mechanism, in which the parameters of the model
are updated on the test input itself. The original formulation trains on a
self-supervised task defined on the test image \cite{sun2020ttt}, and TENT minimises the
entropy of the model's own predictions \cite{wang2021tent}. A recent formulation for
language models goes further and trains the model end to end so that its initialisation
is optimised for the update it will make at test time \cite{sun2025ttte2e}. Interactive segmentation is a
particularly favourable setting for test-time training, because the user supplies
supervision of ground truth quality. Optimisation at click time accordingly has a long
history in interactive segmentation. These approaches range from backpropagating the interaction loss to the inputs or auxiliary latent variables of a frozen network \cite{jang2019brs,sofiiuk2020fbrs}, through updating the model parameters directly from user corrections \cite{kontogianni2020continuous} and, to adapting the model online at inference time in medical imaging \cite{xu2026youpoint,schoen2024adaptingsam}.
These methods share a common property.
The update is a fixed optimisation procedure applied at test time, whose loss, step size
and parameter subset are set by hand and whose objective is to fit the pixels that were
clicked. The training of the model does not control how the rest of the volume responds
to such a step, so whether a correction improves the untouched slices depends on the
region of parameter space that the fine-tuned model occupies.

In this work we propose to train an interactive segmentation model for medical images to
be corrected, so that a click improves not only the slice on which it is placed but the
segmentation of the whole lesion. We refer to this formulation as LeCor (learning to be corrected). We retain the structure of test-time optimisation,
namely a gradient step on a small parameter subset driven by the click, and make the
response of the model to that step a training objective. A small set of adapters termed case adapters is reset at the start of every case and
forms the only part of the model that changes during a session. Their initial values and step sizes are meta-learned so that
one gradient step taken on the clicked slices improves the prediction on the slices that
were not clicked. To our knowledge, this is the first meta-learned test-time training method for
interactive medical image segmentation, in which the update triggered by a correction is
itself learned rather than designed. On 133 held-out lung tumours from five public cohorts LeCor
continues to improve through all seven correction rounds where memory conditioning
saturates, and after the seventh round it stands 4.00 percentage points of Dice (points
hereafter) above the fine-tuned model evaluated with the same seven rounds and 2.82 points
above the same update with a fixed step. Three rounds of LeCor reach the accuracy that the fine-tuned model reaches in seven. Controlled comparisons attribute the gain to
the update itself rather than to the added parameters, and to the meta-learned
initialisation rather than to the learned step sizes. From the standard initialisation every fixed step above 0.01 is harmful, yet
the meta-learned initialisation is updated with benefit at steps three times larger,
so meta-training has not found a better optimiser but has placed the case adapters
where a large step is informative about the rest of the lesion. The approach treats a
correction as a training signal rather than only as a prompt, and the same mechanism
holds potential for other anatomies, imaging modalities and memory based foundation
models.

% =====================================================================================
\section{Related Work}
\label{sec:related}

MedSAM \cite{ma2024medsam} fine-tuned the image encoder and mask decoder of SAM on a
large curated medical corpus, and where such fine-tuning is impractical, parameter
efficient methods dominate. SAMed \cite{zhang2023samed} applies low-rank adaptation
(LoRA) \cite{hu2022lora} to the image encoder, whereas the Medical SAM Adapter
\cite{wu2023medsa} inserts adapter blocks into the encoder and decoder and Conv-LoRA
\cite{zhong2024convlora} adds convolutional structure to the low-rank branch. The video
models have since been adapted to medical images. Ma et al.\ \cite{ma2024sam2benchmark}
benchmarked and fine-tuned SAM~2 across eleven medical modalities, and Medical SAM~2
\cite{zhu2024medicalsam2} and MedSAM2 \cite{ma2025medsam2} treat 2D and 3D medical
segmentation uniformly as video tracking. SLM-SAM~2 \cite{chen2025slmsam2} splits the
memory into short and long term banks to propagate a few annotated slices through a
volume. For SAM~3, an adapter based method has been evaluated on medical images
\cite{chen2025sam3adapter} and a fully fine-tuned medical variant has been proposed
\cite{jiang2026medicalsam3}. Models built
natively for volumes, such as SAM-Med3D \cite{wang2024sammed3d} and SegVol
\cite{du2024segvol}, take the alternative route of a 3D encoder.

The click protocol used by RITM \cite{sofiiuk2022ritm}, FocalClick \cite{chen2022focalclick}
and SimpleClick \cite{liu2023simpleclick} is now standard in the field. It simulates
corrective clicks during training by sampling from the current error map and reports at
evaluation the number of clicks needed to reach a target overlap together with the
number of cases that never reach it. Two lines of work go further and perform an
optimisation at click time. The first optimises inputs or activations. BRS
\cite{jang2019brs} backpropagates the click constraint into the click maps at the network
input, and f-BRS \cite{sofiiuk2020fbrs} moves the optimised variables to auxiliary scale
and bias parameters of intermediate features for speed; both leave the network weights
unchanged, and RITM subsequently argued that better iterative training makes this
refinement unnecessary. The second line optimises weights. Kontogianni et al.\
\cite{kontogianni2020continuous} treat each user correction as a training example and
adapt the model parameters online, both to the current image and across a sequence of
images. In medical imaging, Xu et al.\ \cite{xu2026youpoint} update an interactive model
from clicks during and after each interaction with a loss centred on the clicked
location, in order to handle distribution shift across a sequence of images. Sch\"on et
al.\ \cite{schoen2024adaptingsam} adapt the mask decoder of SAM during usage from the
clicks it receives and its own predictions.

Updating the weights on the test input is the principle of test-time training, which in its original form adapts to each test input
through a self-supervised task \cite{sun2020ttt}. TENT \cite{wang2021tent} minimises the
prediction entropy over target batches, and the recent end-to-end formulation
\cite{sun2025ttte2e} meta-learns the initialisation through the test-time update itself.
On the meta-learning side, MAML \cite{finn2017maml} learns an initialisation from which a
few gradient steps succeed, first-order variants \cite{nichol2018reptile} avoid second
derivatives, and Meta-SGD \cite{li2017metasgd} additionally learns a step size for each
parameter.

% =====================================================================================
\section{Methods}
\label{sec:method}

LeCor comprises an interactive segmentation model for lung tumours obtained by
fine-tuning SAM~3, and a meta-learned test-time update of a small set of case adapters
that is added to it.

\begin{figure}[t]
\centering
\includegraphics[width=\textwidth]{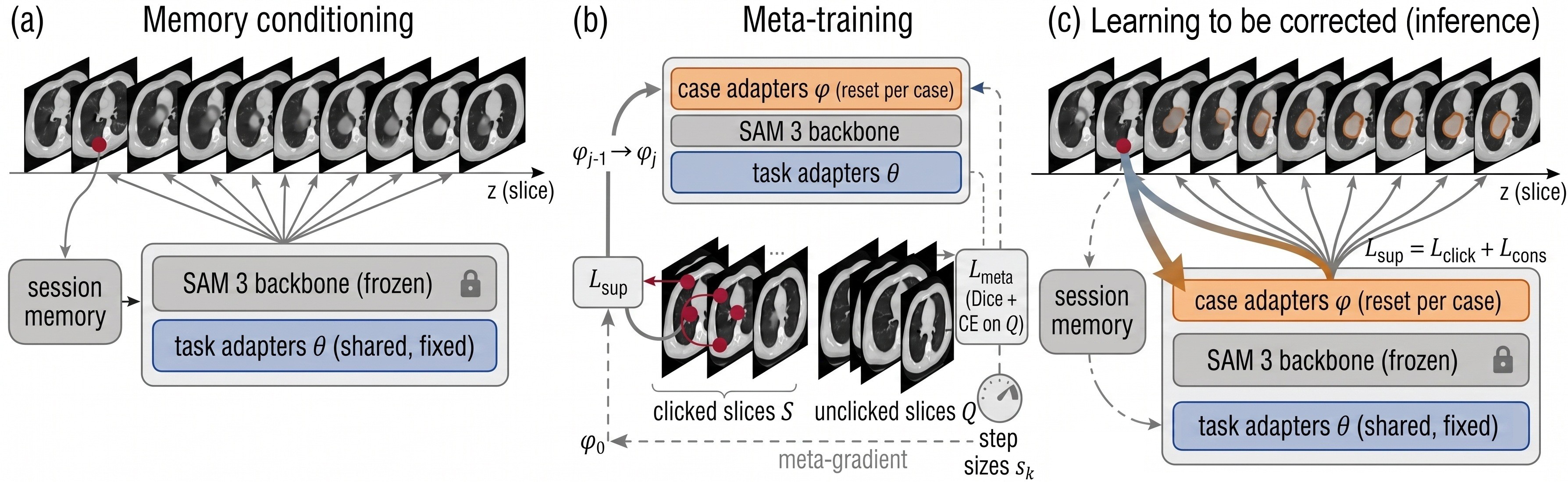}
\caption{Overview. (a) Memory conditioning in the fine-tuned model, in which a click on
one slice is written into the session memory and conditions the slices decoded afterwards
while all weights stay fixed. (b) Meta-training. The inner gradient step uses the clicked
slices $\mathcal{S}$ and the outer objective measures the segmentation
error on the unclicked slices $\mathcal{Q}$ of the same lesion. Both the
initial values $\phi_0$ and the step sizes $s_k$ are learned by minimizing
this outer objective. (c) The LeCor update at inference. Each click drives one gradient step on the case
adapters $\phi$, which are reset at the start of every case. The updated
adapters then refine the slices that were never clicked.}
\label{fig:overview}
\end{figure}

\subsection{Dataset}
\label{sec:data}
The study uses five public CT collections with expert lesion delineations, LUNA25
\cite{peeters2025luna25data,peeters2026luna25} (2,120 patients), NSCLC-Radiomics
\cite{aerts2014nsclc} (421 patients), MSD Task06 Lung \cite{antonelli2022msd} (63
patients), LNDb \cite{pedrosa2021lndb} (166 patients) and 4D-Lung \cite{hugo20174dlung}
(15 patients). A case is
one $z$-contiguous run of slices on which a single lesion is delineated, and the five
collections provide 7,321 cases in total. Volumes are windowed
to $[-1350, 150]$\,HU, mapped to 8 bits and kept at native resolution. The data are
split by patient, 80/10/10 within each collection with a fixed seed, so that all scans
and cases of a patient fall in the same split; this gives 5,880 training, 751 validation
and 690 test cases.

\subsection{Interactive base model}
\label{sec:base}
Each case is presented to the SAM~3 video tracker as a short sequence of slices ordered
along $z$. A single positive point at the centroid of the reference mask on the middle slice of the
lesion starts the session, and the segmentation is propagated forward to the end of the case and
backward to its start, so that the memory of the initial slice conditions both
directions. The backbone is frozen and adapted with low-rank adaptation (LoRA)
\cite{hu2022lora}, which adds a small trainable low-rank correction to selected weight
matrices while the original weights stay unchanged. Adapters of rank 16 on the attention
and MLP projections of the vision encoder, the mask decoder and the memory attention
amount to 2.2\% of the parameters. The model is trained with AdamW on an equally
weighted sum of binary cross entropy and Dice, at the native $1008^2$ resolution of
SAM~3 with at most 64 slices per case. The single point model of Table~\ref{tab:sam3} is trained with light geometric and
intensity augmentation applied with identical parameters to every slice of a case, which
comprises rotation of at most $3^\circ$, isotropic scaling within 0.95 to 1.05,
translation of at most 7 pixels and mild elastic deformation together with brightness,
contrast, gamma, Gaussian noise and blur perturbations without flips. 

Corrective clicks are simulated during training, following the iterative training of
RITM \cite{sofiiuk2022ritm} and the interactive training of SAM~2 \cite{ravi2024sam2}.
In each case at most four slices are corrected, the middle slice and up to three others
drawn uniformly. On a selected slice, each click is drawn uniformly from the false
negative and false positive pixels of the current prediction, positive on a false negative
and negative on a false positive; with probability 0.1 it is drawn from the reference mask
instead, and a slice without error receives a negative click on the background. Each correction is applied before the memory of the slice is encoded, so that it
influences every slice visited afterwards in the same pass, as shown in
Fig.~\ref{fig:overview}a. A correction round is one click on each of the selected slices, so a round consists of
at most four clicks.

\subsection{Learning to be corrected}
\label{sec:ttt}
Let $\mathcal{S}$ be the set of at most four clicked slices of a case and
$\mathcal{Q}$ its remaining, unclicked slices. A session proceeds in rounds
$j=1,\dots,7$. At each round one click is placed on every slice in $\mathcal{S}$, the
model revises its prediction for the whole volume, and accuracy is measured on
$\mathcal{Q}$. Round 0 denotes the state after the initial point before any correction. The model contains three groups of parameters that change on different
timescales. The SAM~3 backbone is frozen. The task adapters $\thetaS$ are the LoRA adapters, which encode lung tumours on CT. They
are shared by all cases and are fixed at inference. The case adapters $\phiF$ are a second, smaller set of LoRA
adapters of rank 4 on the MLP projections of the last eight blocks of the vision encoder
and on the feed-forward layers of the mask decoder, amounting to only 0.09\% of the
backbone parameters. They encode
the current case and are reset at the start of every case. During a session they are the
only parameters that change, by one gradient step after every round of corrections, and
their initial values and step sizes are meta-learned with the model-agnostic
meta-learning framework \cite{finn2017maml} under an objective placed on $\mathcal{Q}$.

A case adapter adds a low-rank term to a frozen linear layer of the network,
\begin{equation}
y \;=\; W x + B A x ,
\label{eq:lora}
\end{equation}
where $W$ is the frozen weight matrix and $A$ and $B$ are the two small matrices of the
adapter. The standard initialisation sets $B$ to zero and draws $A$ at random at the
start of every case. This recovers the fine-tuned model exactly, while the gradient with
respect to $B$ remains non-zero because it is proportional to $Ax$, so the adapters can
learn from the first click. In contrast, Meta-training starts from this initialisation and optimises all tensors of $\phiF_0$

Each case adapter tensor $k$ has its own step size, $s_k = \sigma(a_k)\,g$. The 64
scalars $a_k$ are learned by the outer loop described below, as in Meta-SGD
\cite{li2017metasgd}, and $g$ is a global scale of 0.2 during meta-training and 0.1 at
inference, chosen on the validation split. After meta-training the learned steps range
from 0.015 to 0.050, a factor of three between the most and the least mobile tensors.

At round $j$ the correction is turned into one gradient step on the case adapters,
\begin{equation}
\phiF_j \;=\; \phiF_{j-1} \;-\; s \odot \nabla_{\phiF} L_{\text{sup}}(\phiF_{j-1}),
\qquad
L_{\text{sup}} \;=\; L_{\text{click}} + L_{\text{cons}} ,
\label{eq:inner}
\end{equation}
where $s$ is the vector of step sizes $s_k$ and $\odot$ multiplies each tensor by its
own step size. The loss $L_{\text{sup}}$ has two parts, both computed on the clicked
slices with the current adapters $\phiF_{j-1}$. $L_{\text{click}}$ is the binary cross
entropy between the predicted foreground probability at each clicked pixel and the label
of that click, which is one for a positive click and zero for a negative one. It is the
only term that carries information from the annotator. $L_{\text{cons}}$ is a Dice and
cross entropy loss between the prediction on the clicked slices and a target mask for
those slices. During meta-training the target is the reference mask. At inference no
reference exists, so the target is produced by the model itself. After the clicks of
round $j$ have been written into the session memory, the clicked slices are decoded with
the current adapters and the resulting probability map is binarised at 0.5 to give a
pseudo mask. The session memory is held fixed while the gradient is taken, so the
gradient flows only through the case adapters on the clicked slices. $L_{\text{cons}}$
therefore keeps the parameter update consistent with what the memory has already
inferred from the click, and the update needs nothing beyond what the annotator
supplied. The adapters are not reset between rounds; $\phiF$ accumulates the updates of
a case and is reset only when the next case is opened. After each update the whole volume
is propagated again with the updated adapters, and this propagation is the prediction of
round $j$, as shown in Fig.~\ref{fig:overview}c and summarised in
Algorithm~\ref{alg:inference}. No second derivatives are computed in
Eq.~\ref{eq:inner}, while $\phiF_j$ remains differentiable with respect to the initial
values $\phiF_0$ and to the step parameters $a_k$.

\begin{algorithm}[t]
\caption{Interactive segmentation of one case with LeCor.}
\label{alg:inference}
\begin{algorithmic}[1]
\State $\phiF \gets \phiF_0$ \Comment{case adapters reset; backbone and $\thetaS$ fixed}
\State place the initial point on the middle slice and propagate the volume \Comment{round 0}
\For{$j = 1, \dots, 7$}
  \State the annotator places one click on each slice in $\mathcal{S}$
  \State write the clicks into the session memory and decode the slices in $\mathcal{S}$
  \State $\hat{M} \gets \mathbb{1}\big[p_{\phiF}(\mathcal{S}) > 0.5\big]$ \Comment{pseudo mask, memory held fixed}
  \State $L_{\text{sup}} \gets L_{\text{click}} + L_{\text{cons}}\big(p_{\phiF}(\mathcal{S}), \hat{M}\big)$
  \State $\phiF \gets \phiF - s \odot \nabla_{\phiF} L_{\text{sup}}$ \Comment{Eq.~\ref{eq:inner}; $\phiF$ accumulates across rounds}
  \State propagate the whole volume with the updated $\phiF$ \Comment{prediction of round $j$}
\EndFor
\end{algorithmic}
\end{algorithm}

The training objective measures whether the update helped the slices that were not
clicked,
\begin{equation}
L_{\text{meta}}
 \;=\; \frac{\sum_{j\ge 1} j\, L_{\mathcal{Q}}(\phiF_j)}{\sum_{j\ge 1} j}
   \;+\; \lambda_{\text{a}}\, L_{\mathcal{Q}}(\phiF_0),
\label{eq:outer}
\end{equation}
where $L_{\mathcal{Q}}(\phiF_j)$ is the Dice and cross entropy error on the unclicked
slices after $j$ rounds, evaluated with the reference masks. Later rounds receive a
larger weight, and the second term, with $\lambda_{\text{a}}=0.1$, penalises the initial
adapters from degrading the model before any click. Gradients from Eq.~\ref{eq:outer} update $\phiF_0$, the step
parameters $a_k$ and the task adapters $\thetaS$. The placement of the objective on
$\mathcal{Q}$ rather than on $\mathcal{S}$, as shown in Fig.~\ref{fig:overview}b, is
what distinguishes learning to be corrected from fitting the click, because the error on
the unclicked slices rewards an update that revises the segmentation of the whole lesion
from evidence observed on at most four slices. During meta-training and at evaluation each click is placed at the deepest point of the
current error, the maximum of the distance transform of the deeper of the false negative
and false positive maps. Meta-training runs for three epochs over the 1,478 training cases
that span at least eight slices, with three to four correction rounds per case and six
unclicked slices evaluated per round, using AdamW at a learning rate of $10^{-3}$
($10^{-2}$ for the step parameters) with cosine decay on two A100 GPUs.

\subsection{Evaluation}
\label{sec:setup}
Test-time training needs unclicked slices to be evaluated on, so the evaluation uses the
cases that span at least eight slices, 176 validation and 133 test cases drawn from all
five collections. Each session comprises seven correction rounds on the middle slice and up to three
further slices drawn once per case, fixed for all rounds and all experiments, with the
click rule of the meta-training. Accuracy is reported as Dice on the unclicked slices unless stated otherwise.

Six experiments listed in Table~\ref{tab:ablation} separate the ingredients of the
method. The
first two experiments start from the fine-tuned model with the standard initial values of
the case adapters. In the no-update experiment the case adapters remain at these values
and the corrections act through the session memory only, which corresponds to the
fine-tuned model without the LeCor update. In the fixed-step experiment the case
adapters are updated with Eq.~\ref{eq:inner} at a single step of $10^{-2}$, the
best value in a sweep on the validation cases. The remaining four experiments use the meta-trained
checkpoint. One keeps the meta-learned initial values but never runs the update, which
tests whether the meta-learned parameters help on their own. Two others swap one
ingredient at a time, driving the standard initial values with the learned steps or the
meta-learned initial values with the fixed step of $10^{-2}$. The last experiment is LeCor, the meta-learned initial values with the learned
steps and the update enabled. All
experiments are run on the same 133 cases with the same seed. The two
experiments without meta-learning are evaluated on the original fine-tuned weights and the
other four on the meta-trained checkpoint.

All metrics are computed on the volume formed by stacking the per-slice predictions and
references of a case. They comprise Dice, IoU, precision, recall, the 95th percentile
Hausdorff distance (HD95) and the normalised surface Dice (NSD)
\cite{nikolov2021deepmind}. Surface distances are given in voxels because the images are
not resampled. Annotation cost
is reported as the mean number of correction rounds needed to reach a Dice target (NoC)
and the number of cases that never reach it within seven rounds (NoF), with failures
charged the full budget. Differences between experiments are computed per case and summarised
by their mean, a 95\% confidence interval from 10,000 bootstrap resamples and a two
sided Wilcoxon signed-rank test.

% =====================================================================================
\section{Results}
\label{sec:results}

\subsection{Interactive base model}
\label{sec:res:base}
The accuracy of SAM~3 on lung tumours before any test-time update is measured on the
full 690-case test set, as shown in Table~\ref{tab:sam3}. Without adaptation SAM~3 locates the lesion but
grossly over-segments it, and fine-tuning with LoRA raises the Dice obtained from a
single point prompt from 0.298 to 0.757. Training with simulated corrections raises the accuracy reached
after seven rounds of clicks to 0.765 and reduces HD95 from 7.6 to 2.7 voxels, so the
interactive model is the base for all subsequent experiments. Over the seven rounds
precision rises from 0.78 to 0.85 while recall stays at 0.71.

\begin{table}[t]
\caption{SAM~3 adaptation on the 690-case 3D test set. HD95 is given in voxels and NSD
at a tolerance of 2 voxels. Fine-tuned rows train the encoder, decoder and memory
adapters. Best value per column in bold; recall is highest for the zero-shot model
because it over-segments.}
\label{tab:sam3}
\centering
\footnotesize
\begin{tabular}{@{}lrrrrr@{}}
\toprule
Model & Dice & Prec. & Rec. & HD95$\downarrow$ & NSD@2 \\
\midrule
SAM~3 zero-shot, 1 point                   & 0.2978 & 0.2437 & \textbf{0.9414} & 100.15 & 0.3634 \\
LoRA fine-tuned, 1 point                   & 0.7569 & 0.7622 & 0.7825 & 5.87 & 0.9468 \\
Interactive model, 1 point                 & 0.7237 & 0.7789 & 0.7058 & 7.60 & 0.9323 \\
Interactive model, 7 rounds                & \textbf{0.7649} & \textbf{0.8494} & 0.7055 & \textbf{2.67} & \textbf{0.9754} \\
\bottomrule
\end{tabular}
\end{table}

\subsection{Accuracy across correction rounds}
\label{sec:res:main}
\begin{figure}[t]
\centering
% Fig. 2: pgfplots groupplot from the per-round means. (a) fig_curves_ci.csv: no update and
% fixed step on the original fine-tuned weights, the other two on the meta-trained checkpoint.
% (b) fig_attr_ci.csv + fig_attr_c4lr_ci.csv: one evaluation pass on the meta-trained checkpoint.
\begin{tikzpicture}
\begin{groupplot}[
  group style={group size=2 by 1, horizontal sep=10pt, ylabels at=edge left, yticklabels at=edge left},
  width=0.48\textwidth, height=0.4\textwidth,
  xlabel={Correction round}, ylabel={Dice on unclicked slices},
  xmin=-0.2, xmax=7.2, ymin=0.74, ymax=0.86,
  xtick={0,1,2,3,4,5,6,7}, ytick={0.74,0.76,0.78,0.80,0.82,0.84,0.86},
  yticklabel style={/pgf/number format/fixed, /pgf/number format/precision=2},
  grid=major, grid style={gray!20},
  title style={font=\footnotesize, yshift=-2pt},
  tick label style={font=\footnotesize}, label style={font=\small},
  every axis plot/.append style={line width=0.9pt, mark size=1.6pt}]
\nextgroupplot[title={(a) Update mechanism},
  legend to name=leg:curves, legend columns=2, legend cell align=left,
  legend style={font=\tiny, draw=none, fill=none,
    /tikz/every even column/.append style={column sep=10pt}},
  legend image post style={xscale=0.5}]
\addplot[color=black!45, mark=o, forget plot] coordinates {(0,0.7520)(1,0.7668)(2,0.7801)(3,0.7860)(4,0.7865)(5,0.7883)(6,0.7914)(7,0.7869)};
\addplot[color=blue!70!black, mark=square, forget plot] coordinates {(0,0.7520)(1,0.7645)(2,0.7797)(3,0.7827)(4,0.7825)(5,0.7916)(6,0.7956)(7,0.7987)};
\addplot[color=green!55!black, mark=triangle, dashed, forget plot] coordinates {(0,0.7487)(1,0.7696)(2,0.7720)(3,0.7704)(4,0.7727)(5,0.7707)(6,0.7682)(7,0.7658)};
\addplot[color=orange!90!black, mark=*, forget plot] coordinates {(0,0.7487)(1,0.7860)(2,0.7982)(3,0.8068)(4,0.8142)(5,0.8119)(6,0.8208)(7,0.8269)};
\addlegendimage{color=black!45, mark=o, line width=0.9pt, mark size=1.6pt}
\addlegendentry{Fine-tuned, standard init., no update}
\addlegendimage{color=blue!70!black, mark=square, line width=0.9pt, mark size=1.6pt}
\addlegendentry{Fine-tuned, standard init., fixed step}
\addlegendimage{color=brown!80!black, mark=pentagon, line width=0.9pt, mark size=1.6pt}
\addlegendentry{Meta-trained, standard init., fixed step}
\addlegendimage{color=violet!80!black, mark=diamond, line width=0.9pt, mark size=1.6pt}
\addlegendentry{Meta-trained, standard init., learned steps}
\addlegendimage{color=green!55!black, mark=triangle, dashed, line width=0.9pt, mark size=1.6pt}
\addlegendentry{Meta-trained, meta init., no update}
\addlegendimage{color=teal!80!black, mark=triangle*, line width=0.9pt, mark size=1.6pt}
\addlegendentry{Meta-trained, meta init., fixed step}
\addlegendimage{color=orange!90!black, mark=*, line width=0.9pt, mark size=1.6pt}
\addlegendentry{Meta-trained, meta init., LeCor}
\nextgroupplot[title={(b) Initialisation and step rule}]
\addplot[color=brown!80!black, mark=pentagon, forget plot] coordinates {(0,0.7600)(1,0.7762)(2,0.7883)(3,0.7861)(4,0.7903)(5,0.7908)(6,0.7903)(7,0.7863)};
\addplot[color=violet!80!black, mark=diamond, forget plot] coordinates {(0,0.7600)(1,0.7765)(2,0.7841)(3,0.7811)(4,0.7794)(5,0.7677)(6,0.7756)(7,0.7739)};
\addplot[color=teal!80!black, mark=triangle*, forget plot] coordinates {(0,0.7487)(1,0.7837)(2,0.7938)(3,0.7941)(4,0.8005)(5,0.8044)(6,0.8081)(7,0.8087)};
\addplot[color=orange!90!black, mark=*, forget plot] coordinates {(0,0.7487)(1,0.7860)(2,0.7982)(3,0.8068)(4,0.8142)(5,0.8119)(6,0.8208)(7,0.8269)};
\end{groupplot}
\end{tikzpicture}\\[2pt]\ref{leg:curves}
\caption{Dice on the unclicked slices after each correction round, mean over the 133 test
cases. Each legend entry names the task adapters, the initial values of the case adapters
and the step rule, the three columns by which Table~\ref{tab:ablation} defines an
experiment. (a) What a correction contributes, from the memory alone to LeCor. (b) The
four combinations of initial values and step rule, all on the meta-trained task adapters.}
\label{fig:curves}
\end{figure}
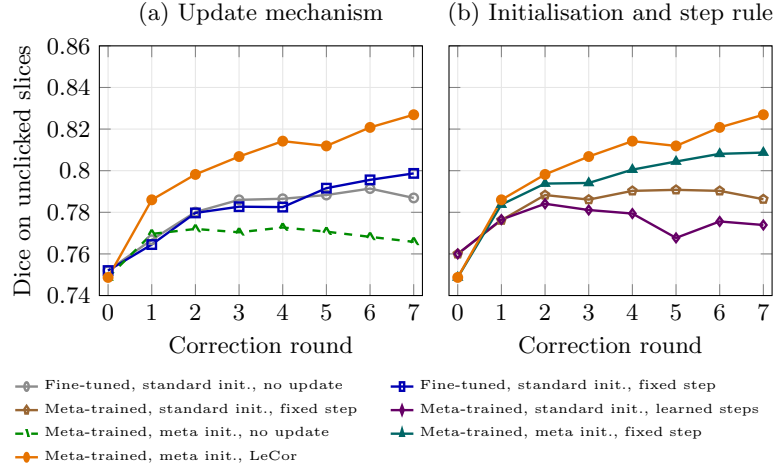

Figure~\ref{fig:curves}a shows Dice on the unclicked slices of the 133 test cases after
each correction round, where a round is one click on each of at most four slices. Without any update the fine-tuned model improves with clicks, since prompting
through the memory remains effective, but it saturates at a Dice of 0.791 by round 6 and
declines at round 7. The fixed-step update with the form of test-time optimisation used in prior work also
improves the segmentation, but only in later rounds. It makes no difference or is
slightly harmful for four rounds before turning positive, and it ends 1.18 points above no
update. As listed in Table~\ref{tab:ablation}, LeCor improves the segmentation from the first
round onwards and is still rising at round 7, where it reaches a Dice of 0.827, 4.00
points above no update. The margin is 3.55 points
when each experiment is compared at its own best round and 2.19 points when averaged
over all rounds, and the three summaries decrease in that order because the advantage
accumulates with interaction. At round 0 the meta-learned initial values lie 0.33 points
below the no update experiment, so the advantage is produced entirely by the updates. Against the fixed-step update the margin at round 7 is 2.82 points, so meta-learning roughly triples the final effect of the same update
mechanism. All differences quoted in this section are significant at $p<0.001$ by the
Wilcoxon signed-rank test unless stated otherwise.

\subsection{Qualitative analysis}
\label{sec:res:qual}
\begin{figure}[h!]
\centering
\includegraphics[width=0.8\textwidth]{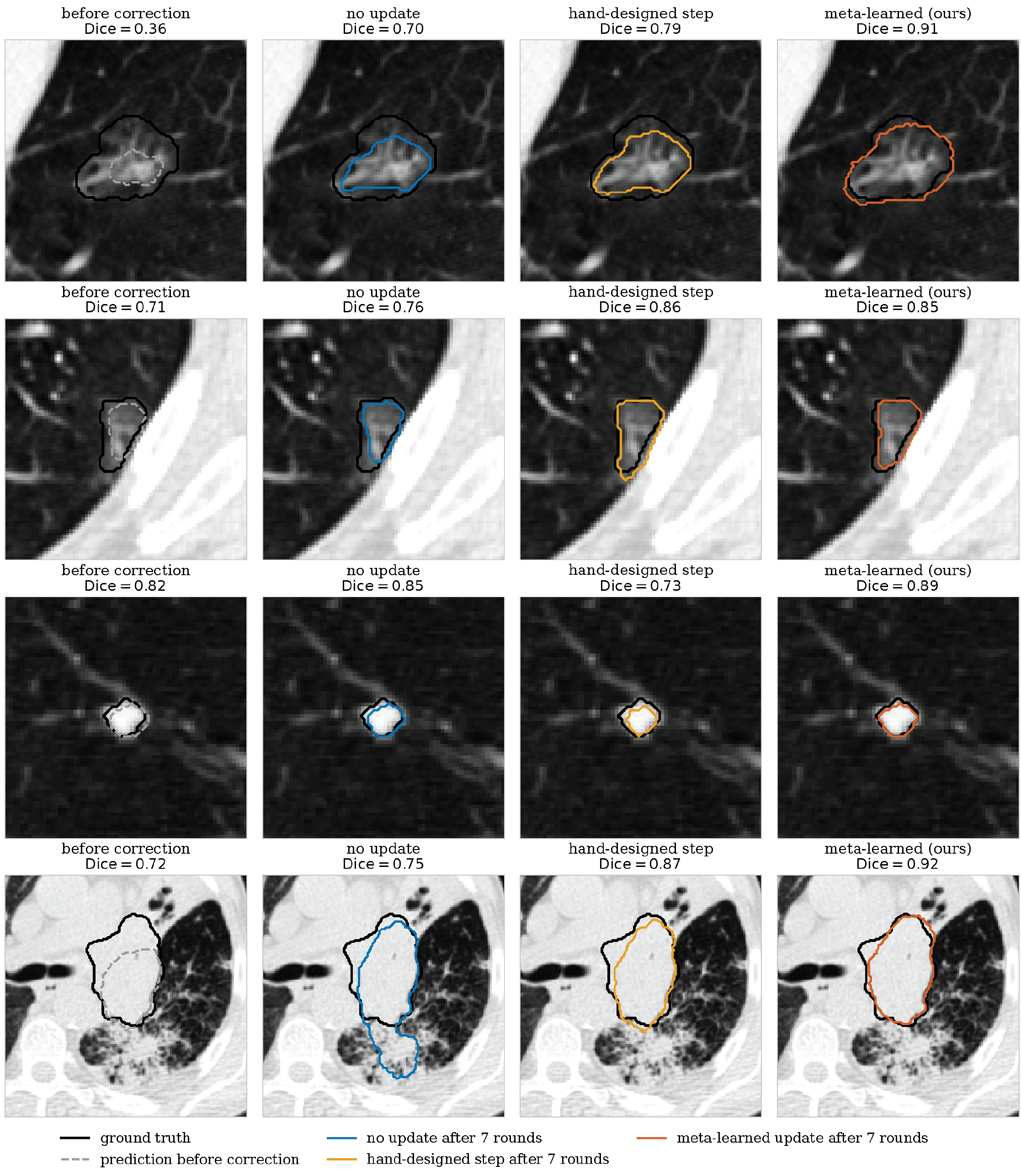}
\caption{Four test cases, each shown on a slice that received no click. The columns are
the state before any correction and, after seven correction rounds, the fine-tuned task
adapters with standard initial values and no update, the same adapters with the
fixed-step update, and LeCor. Black is the reference contour, grey dashed the prediction
before any correction, and the coloured contour the state after seven rounds. Panel
titles give Dice on the slice shown.}
\label{fig:qual}
\end{figure}

Figure~\ref{fig:qual} shows four test cases on a slice that received no click, before any
correction and after seven rounds without update, with the fixed-step update and with
LeCor. In the first case the initial prediction covers only part of the lesion,
with a Dice of 0.36 on the slice shown; seven rounds of clicks on other slices recover
about half of it through the memory alone, the fixed step recovers more, and LeCor traces
the whole lesion, with Dice values of 0.70, 0.79 and 0.91 respectively. In
the second case the lesion abuts the chest wall, and both updates correct the pleural side
of the contour to a similar degree on this slice, whereas over the unclicked slices of the
whole case LeCor remains ahead, at 0.85 against 0.80. In the third case the
fixed-step update degrades a small nodule that the model without update had segmented
almost correctly, which illustrates the late and unreliable behaviour of the fixed step,
while LeCor improves the same contour. In the fourth case the fine-tuned
model leaks out of the tumour into the consolidated lung tissue below it, and seven rounds
of clicks on other slices leave the leak in place because the weights that produced it
never change, whereas LeCor removes it.

\begin{table}[t]
\caption{Ablation of LeCor on the 133 test cases. Each row is one experiment, defined by
the task adapters it uses and by whether the case adapters start from the meta-learned
initial values, whether they are updated at test time and with which step rule; the last
row is LeCor. Dice is the mean on the unclicked slices at round 7; $\Delta$ is the
per-case difference to no update at round 7 in percentage points ($^{*}$ $p<0.05$,
$^{**}$ $p<0.001$, Wilcoxon signed-rank test); NoF and NoC are the number of cases that
never reach $\dice\ge0.80$ within seven rounds and the mean number of rounds needed, with
failures charged seven. Best value per column in bold.}
\label{tab:ablation}
\centering
\footnotesize
\setlength{\tabcolsep}{4pt}
\begin{tabular}{@{}lccc@{\hspace{10pt}}rr@{\hspace{10pt}}rr@{}}
\toprule
& \multicolumn{3}{c}{Case adapters} & & & \multicolumn{2}{c@{}}{$\dice\ge0.80$} \\
\cmidrule(lr){2-4}\cmidrule(l){7-8}
Task adapters & Meta init. & Update & Step & Dice & $\Delta$ & NoF & NoC \\
\midrule
fine-tuned   & --         & --         & --      & 0.787 & $\text{--}\phantom{^{**}}$   & 47 & 2.72 \\
fine-tuned   & --         & \checkmark & fixed   & 0.799 & $+1.2^{**}$                   & 43 & 2.65 \\
meta-trained & \checkmark & --         & --      & 0.766 & $-2.1^{*}\phantom{^{*}}$      & 59 & 3.33 \\
meta-trained & --         & \checkmark & learned & 0.774 & $-1.3\phantom{^{**}}$         & 42 & 2.87 \\
meta-trained & \checkmark & \checkmark & fixed   & 0.809 & $+2.2^{*}\phantom{^{*}}$      & 32 & 2.56 \\
meta-trained & \checkmark & \checkmark & learned & \textbf{0.827} & $\mathbf{+4.0}^{**}$ & \textbf{27} & \textbf{2.17} \\
\bottomrule
\end{tabular}
\end{table}

\subsection{Source of the gain}
\label{sec:res:attr}
LeCor differs from the fine-tuned model in three respects. It carries
additional parameters, which start from meta-learned values and are updated at test time
with learned step sizes. The contribution of each is separated below. The additional
parameters alone do not produce the gain. The experiment that retains the meta-learned initial values without the update is identical to LeCor in all other respects. As shown in Table~\ref{tab:ablation}, it falls 2.11 points below no update at round 7 and 3.44 points below LeCor when averaged over all rounds. The
meta-learned parameters are therefore of value only as a starting point for correction.

The remaining two ingredients are separated by the four combinations of initial values
and step rule in
Fig.~\ref{fig:curves}b, all measured on the meta-trained checkpoint, in which the standard
initial values with the fixed step read 0.7863. The no-update
experiment evaluated on the meta-trained checkpoint differs from the one on the original
fine-tuned weights by 0.37 points ($p=0.81$), so the joint update of the task adapters
did not move them measurably. At the fixed step, the meta-learned initial values improve Dice by 2.24 points
over the standard ones ($p<0.001$). On the meta-learned initial values, the learned steps improve Dice by 1.82 points over
the fixed step. The same learned steps applied to
the standard initial values are harmful, finishing 1.25 points below the fixed step and
below no update, and a sweep of the fixed step on the 176 validation cases explains this
behaviour. From the standard initial values every fixed step of
0.02 or larger is worse than not updating at all and 0.01 gains only 0.45 points, so the
steps that benefit
the meta-learned initial values, 0.015 to 0.05, are harmful when applied to the standard
ones. The
contribution of meta-learning is therefore the starting point of the update.

\subsection{Annotation cost}
\label{sec:res:noc}
\begin{figure}[t]
\centering
% Fig. 4: pgfplots from the values of Table~\ref{tab:noc}. Shared legend below the panels
% (legend to name; needs two compilation passes).
\begin{tikzpicture}
\begin{axis}[
  width=0.5\textwidth, height=0.36\textwidth, ybar=5pt, bar width=10pt,
  symbolic x coords={0.75,0.80,0.85}, xtick=data,
  xlabel={Dice target}, ylabel={Cases reaching the target}, ymin=0, ymax=140,
  enlarge x limits=0.3, tick label style={font=\footnotesize}, label style={font=\small},
  nodes near coords, every node near coord/.append style={font=\tiny, yshift=-1pt},
  legend to name=leg:noc, legend columns=2, legend style={font=\footnotesize, draw=none, fill=none, /tikz/every even column/.append style={column sep=8pt}},]
\addplot[fill=black!35, draw=black!50] coordinates {(0.75,98)(0.80,86)(0.85,58)};
\addplot[fill=orange!85!black, draw=orange!60!black] coordinates {(0.75,120)(0.80,106)(0.85,69)};
\legend{No update, LeCor}
\end{axis}
\end{tikzpicture}\hfill
\begin{tikzpicture}
\begin{axis}[
  width=0.5\textwidth, height=0.36\textwidth, ybar=5pt, bar width=10pt,
  symbolic x coords={0.75,0.80,0.85}, xtick=data,
  xlabel={Dice target}, ylabel={Mean correction rounds needed}, ymin=0, ymax=5.2,
  enlarge x limits=0.3, tick label style={font=\footnotesize}, label style={font=\small},
  nodes near coords, every node near coord/.append style={font=\tiny, yshift=-1pt},
  ]
\addplot[fill=black!35, draw=black!50] coordinates {(0.75,1.95)(0.80,2.72)(0.85,4.18)};
\addplot[fill=orange!85!black, draw=orange!60!black] coordinates {(0.75,1.28)(0.80,2.17)(0.85,4.14)};

\end{axis}
\end{tikzpicture}
\\[2pt]\ref{leg:noc}
\caption{Successes and annotation cost at three Dice targets on the 133 test cases for the
no-update experiment and LeCor. Left, the number of the 133 cases that reach
the target within seven rounds; right, the mean number of correction rounds needed, with
failures charged seven.}
\label{fig:noc}
\end{figure}
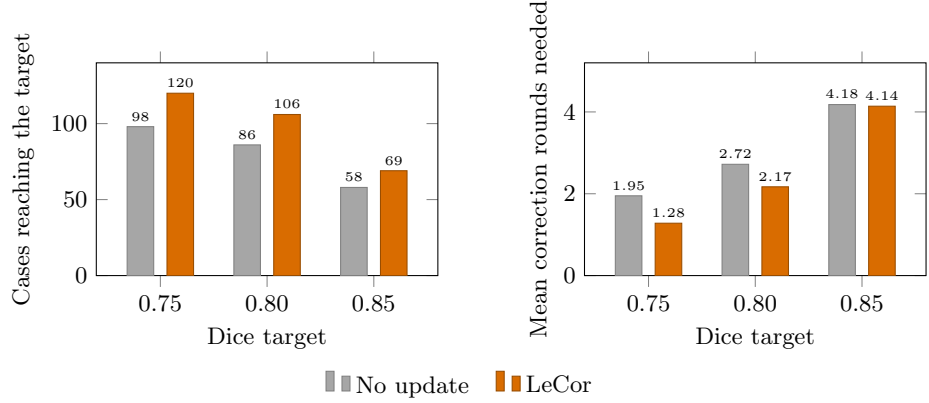

The gain in Dice translates into annotation cost, measured as the number of correction
rounds a case needs and the number of cases that never reach a given contour quality, as
shown in Table~\ref{tab:noc} and Fig.~\ref{fig:noc}. At a target of $\dice\ge0.80$ the number of
cases that never reach the target within seven rounds falls from 47 with no update to 27
with LeCor, and the mean number of rounds needed falls by 0.56. At
$\dice\ge0.75$ the failures fall by 63\%, and at $\dice\ge0.85$ failures still convert
into successes, from 75 to 64, while the saving in rounds is no longer significant. The
benefit is therefore concentrated in reaching moderate contour quality faster. The fixed-step update leaves the failure count essentially unchanged (43 against 47),
the meta-learned initial values without the update increase it to 59, and the
meta-learned initial values with the fixed step recover most of the reduction (32), with the learned steps
contributing the remainder. The best Dice reached by the no-update experiment within
seven rounds is 0.791. LeCor exceeds this value significantly from round 3
onwards (1.54 points, $p<0.01$), so three rounds of LeCor achieve what
seven rounds of memory conditioning achieve. The improvement
is broad rather than driven by outliers. At round 7 the method improves 96 of the 133
cases, by 6.59 points on average, and degrades 37, by 2.73 points on average. The number
of cases in which clicking leaves the contour worse than it was before any correction
falls from 38 to 13.

\begin{table}[t]
\caption{Annotation cost on the 133 test cases at three Dice targets. Each row names the
task adapters, the initial values of the case adapters and the step rule, as in
Table~\ref{tab:ablation} and in the legends of Fig.~\ref{fig:curves}. NoF is the number
of cases that never reach the target within seven rounds and NoC the mean number of rounds
needed, with failures charged seven. For LeCor the reduction in rounds
against no update is 0.67 rounds at $\dice\ge0.75$ ($p<0.01$), 0.56 at $\dice\ge0.80$
($p<0.05$) and not significant at $\dice\ge0.85$. Best value per column in bold.}
\label{tab:noc}
\centering
\footnotesize
\setlength{\tabcolsep}{4.5pt}
\begin{tabular}{@{}lrrrrrr@{}}
\toprule
& \multicolumn{2}{c}{$\dice\ge0.75$} & \multicolumn{2}{c}{$\dice\ge0.80$} & \multicolumn{2}{c@{}}{$\dice\ge0.85$} \\
\cmidrule(lr){2-3}\cmidrule(lr){4-5}\cmidrule(l){6-7}
Experiment & NoF & NoC & NoF & NoC & NoF & NoC \\
\midrule
Fine-tuned, standard init., no update & 35 & 1.95 & 47 & 2.72 & 75 & 4.18 \\
Fine-tuned, standard init., fixed step & 30 & 1.91 & 43 & 2.65 & 68 & 4.09 \\
Meta-trained, meta init., no update & 33 & 2.02 & 59 & 3.33 & 89 & 4.81 \\
Meta-trained, standard init., learned steps & 17 & 1.32 & 42 & 2.87 & 76 & 4.41 \\
Meta-trained, meta init., fixed step & 17 & 1.34 & 32 & 2.56 & 73 & 4.39 \\
Meta-trained, meta init., LeCor & \textbf{13} & \textbf{1.28} & \textbf{27} & \textbf{2.17} & \textbf{64} & 4.14 \\
\bottomrule
\end{tabular}
\end{table}

\subsection{Computational cost and portability}
\label{sec:res:cost}
On a single A100 the no-update experiment takes 17.7\,s per case and LeCor
24.6\,s, an increase of 0.99\,s per correction round. The increase corresponds to the
backward pass that each round adds. The two updating experiments cost 24.6 to 24.8\,s and
the two non-updating experiments 17.6 to 17.7\,s, so the cost lies in the backward pass
and not in the meta-learned parameters. However, the meta-learned initial values and step sizes are tied to the task adapters
they were trained with. Placed on the original
fine-tuned weights instead of the meta-trained checkpoint, the same initial values and
step sizes reduce Dice to 0.503 at the first update, after which the full seven rounds
are needed to recover, and with the update disabled the
meta-learned initial values are below no update, as shown in Table~\ref{tab:ablation}. The case
adapters and the task adapters therefore form one checkpoint that should be released
together and deployed with the update enabled.

% =====================================================================================
\section{Discussion}
\label{sec:discussion}

The control experiments locate the gain precisely. Writing corrections into the model
with a fixed, hand designed rule improves the segmentation, but the improvement arrives
late and only within a narrow band of step sizes. Meta-learning the rule makes the same
mechanism effective from the first correction, and the comparison of initial values
against step rules attributes this to the starting point of the update rather than to
the step rule. A step size that is harmful from the standard starting point is
beneficial from the meta-learned one. Meta-training has therefore not produced a better
optimiser but has moved the case adapters to a region of parameter space in which a
gradient step that is harmful elsewhere is instead informative about the rest of the
lesion. Placing the same starting point and steps on the original fine-tuned weights supports
the same interpretation, since there they are harmful on the first update.

For the annotator the benefit is that a contour of moderate quality is reached in fewer
correction rounds and that cases which never reached a given quality within the click
budget now do so at every target considered, while the proportion of cases in which the
corrections degrade the segmentation, leaving the final contour less accurate than the
initial prediction, is reduced substantially. These gains come at the additional cost of
one backward pass per round. The method requires only a
model that processes a volume under a shared memory, so that the effect of a correction
on the untouched slices is measurable, and a small set of parameters that can be reset
per case. The same mechanism therefore holds potential for other anatomies, modalities and memory
based foundation models, and for any interactive annotation workflow in which corrections
are currently consumed as prompts.

The present results are established for lesions spanning at least eight slices, since
the meta-objective requires unclicked slices. Drawing the
objective from neighbouring cases of the same patient would extend the method to small
nodules. The simulated clicks always target the deepest error of the current prediction, whereas
an annotator may click elsewhere. How the update responds to clicks that do not target
the deepest error remains to be measured, and a reader study with clicks chosen by
clinicians is therefore the natural next step. The placement of the objective, the weighting of its terms and the joint update of the
task adapters were fixed by design, so further gains from tuning them remain available. The reduction in rounds is established at a $\dice\ge0.80$ target and not at
$\dice\ge0.85$, where failures still fall. The method currently shortens the interaction
needed to reach a good contour rather than the interaction needed to reach an excellent
one.

% =====================================================================================
\section{Conclusion}
\label{sec:conclusion}

We have shown with LeCor that an interactive segmentation model for lung tumours can be
trained to learn from its corrections by a meta-learned update of a small set of case
adapters. LeCor gains 4.00 points of Dice over the fine-tuned model at round 7 and 2.82 points over the same
update with a fixed step, cuts failures at a $\dice\ge0.80$ target by 43\% and reaches in
three rounds what memory conditioning reaches in seven, for one second of GPU time per
round. The gain is produced by the test-time update itself and is attributable to the
meta-learned starting point from which the update begins.

\subsubsection*{Code and models}
Training, evaluation and analysis code, together with the scripts that regenerate every
number and figure in this paper from the per-case outputs, will be released on
acceptance.

\bibliographystyle{splncs04}
\bibliography{refs}

\end{document}